\documentclass{article}
\usepackage{ijcai19}

\usepackage{times}
\usepackage{soul}
\usepackage{url}
\usepackage[hidelinks]{hyperref}
\usepackage[utf8]{inputenc}
\usepackage[small]{caption}
\usepackage{graphicx}
\usepackage{amsmath}
\usepackage{booktabs}
\usepackage{algorithm}
\usepackage{algorithmic}
\usepackage{amsmath}
\usepackage{amsfonts}
\usepackage{breqn}
\usepackage{enumerate}
\usepackage{multirow}
\usepackage{array}
\usepackage{subfigure}
\usepackage{color}
\usepackage{graphicx}  

\title{Hallucinated Adversarial Learning for Robust Visual Tracking}

\author{
    Qiangqiang Wu$^{1}$, Zhihui Chen$^{1}$, Lin Cheng$^{1}$, Yan Yan$^{1}$, Bo Li$^{2}$, Hanzi Wang$^{1*}$
    \affiliations
    $^{1~}$Department of Computer Science, Xiamen University, Xiamen, China
    \affiliations
    $^{2~}$Department of Computer Science and Engineering, Beihang University, Beijing, China \emails
    qiangwu@stu.xmu.edu.cn, \{yanyan, hanzi.wang\}@xmu.edu.cn, \\zhihui.qz.chen@gmail.com, cheng.charm.lin@hotmail.com, libo@buaa.edu.cn
}

\begin{document}

\maketitle

\begin{abstract}


Humans can easily learn new concepts from just a single exemplar, mainly due to their remarkable ability to imagine or hallucinate what the unseen exemplar may look like in different settings. Incorporating such an ability to hallucinate diverse new samples of the tracked instance can help the trackers alleviate the over-fitting problem in the low-data tracking regime. To achieve this, we propose an effective adversarial approach, denoted as adversarial ``hallucinator" (AH), for robust visual tracking. The proposed AH is designed to firstly learn transferable non-linear deformations between a pair of same-identity instances, and then apply these deformations to an unseen tracked instance in order to generate diverse positive training samples. By incorporating AH into an online tracking-by-detection framework, we propose the hallucinated adversarial tracker (HAT), which jointly optimizes AH with an online classifier (e.g., MDNet) in an end-to-end manner. In addition, a novel selective deformation transfer (SDT) method is presented to better select the deformations which are more suitable for transfer. Extensive experiments on 3 popular benchmarks demonstrate that our HAT achieves the state-of-the-art performance.

\end{abstract}

\section{Introduction}
Given the initial state of a target at the first frame, generic visual tracking aims to estimate the trajectory of the target at subsequent frames in a video.
Despite the outstanding success achieved by deep convolutional neural networks (CNNs) in a variety of computer vision tasks \cite{resnet,hallucinator}, their impact in visual tracking is still limited. The main reason is that deep CNNs greatly rely on the large-scale annotated training data. For online tracking, it is impossible to gather enough training data since the tracker is required to track arbitrary objects. Thus the problem of learning an effective CNN model for visual tracking is particularly challenging, mainly due to limited online training data.






To alleviate this problem, one strategy is to treat visual tracking as a more general similarity learning problem, thus enabling deep CNNs (e.g., Siamese networks) to be trained with large-scale annotated datasets in an offline manner. However, these Siamese network based trackers \cite{SiamFC} still cannot achieve high accuracies on the benchmarks, since they inherently lack the online adaptability. 
Another strategy is to effectively leverage few online training samples and adopt the online learning based tracking-by-detection schema. This schema based trackers maintain an online CNN classifier, which models the temporal variations of the target appearance by updating the network parameters. Compared with the Siamese network based trackers, they gain a large accuracy improvement. However, they may easily suffer from the over-fitting problem due to the limited online training samples (especially for the positive training samples), thus leading to suboptimal tracking performance. 
\setlength{\belowcaptionskip}{-0.25cm} 
 \begin{figure}
\begin{center}
   \includegraphics[width=1.00\linewidth]{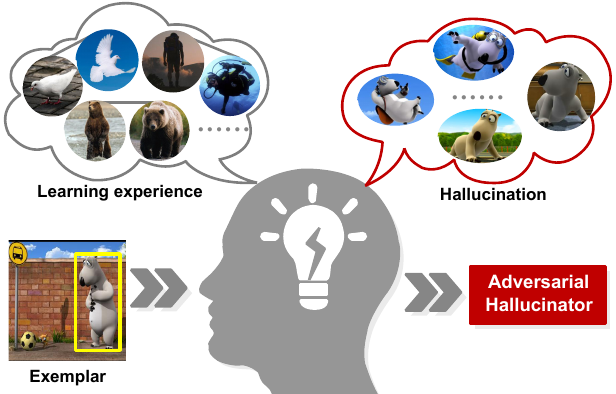} 
\end{center}
\vspace{-0.4cm}
 \caption{Given an unseen exemplar (a bear), humans can effectively hallucinate what the bear may look like in different views or poses based on their previous learning experience. We propose an adversarial hallucinator, which mimics such hallucination or imagination for robust visual tracking. }
\label{pipeline_first}
\end{figure}

Compared with state-of-the-art CNN-based trackers, visual tracking is a relatively simple task for humans. Although how human brain works is far from being fully understood, one can conjecture that humans have a remarkable imaginary mechanism derived from their previous learning experience. That is, as illustrated in Fig. \ref{pipeline_first}, humans can firstly learn many shared modes of variation (e.g., rotation, illumination change and translation) from different pairs of same-identity instances. Then they are able to hallucinate what novel instances look like in different surroundings or poses, by applying their previous learned modes of variation to novel instances. For example, we can learn the motion of rotation from a windmill. Based on it, we can easily imagine how a completely different windmill or even an electric fan rotates. Interestingly, it seems that we build a visual classifier in the brain and then hallucinate novel deformable samples of the exemplar to train the classifier, which is particularly similar to the data augmentation technique in machine learning.
 
In this paper, our main motivation is to help CNN-based trackers do such ``imagination" or ``hallucination", so that they can achieve robust tracking in the low-data tracking regime. To achieve this, we propose a novel adversarial ``hallucinator" (AH), which is based on an encoder-decoder generative adversarial network. The proposed AH learns non-linear deformations using different instance pairs collected from large-scale datasets, and then AH can effectively generate new deformable samples of an unseen instance $x$ by applying the learned deformations to $x$. For the offline training of AH, we present a deformation reconstruction loss, which enables AH to be trained in a self-supervised manner without the need of mining additional samples.
Based on AH, we further propose HAT (hallucinated adversarial tracker), which incorporates AH into a tracking-by-detection framework. Specifically, our AH is jointly optimized with the online CNN classifier (MDNet \cite{MDNet}) in an end-to-end manner, which can effectively help the CNN classifier alleviate the over-fitting problem, thus leading to better generalization of the tracker. In addition, we present a novel selective deformation transfer (SDT) method to effectively select the suitable deformations for better transfer. To sum up, this paper has the following contributions:
\begin{itemize}
\item We propose an adversarial hallucination method, namely adversarial ``hallucinator" (AH), which mimics the human imaginary mechanism for data augmentation. To effectively train AH in a self-supervised manner, a novel deformation reconstruction loss is presented.
\item Based on AH, we present a hallucinated adversarial tracker (HAT), which jointly optimizes our AH and the online classifier (MDNet) in an end-to-end manner. The joint learning schema can effectively help the online classifier alleviate the over-fitting problem.
\item We propose a novel selective deformation transfer (SDT) method to better select the deformations which are more suitable for transfer. 
\end{itemize}

We perform comprehensive experiments on three popular benchmarks: OTB-2013 \cite{OTB50}, OTB-2015 \cite{OTB100}, and VOT-2016 \cite{VOT2016}. Experimental results demonstrate that our HAT performs favorably against state-of-the-art trackers. In particularly, HAT achieves the leading accuracy (95.1\%) on OTB-2013.


\section{Related Work} 
\textbf{CNN-based tracking.} The CNN-based trackers typically employ an one-stage template matching framework or a two-stage classification framework. The representative one-stage framework is based on deep Siamese networks. Starting from SiamFC \cite{SiamFC}, many efforts have been made, including target template modeling \cite{memtrack}, proposal generation \cite{siamrpn}, network architecture design \cite{detailsiam}, and loss function design \cite{multisiam}. In comparison, the two-stage framework consists of two steps: 1) Sample target candidates around the estimated location in the previous frame. 2) Classify all the candidates to obtain the best one in terms of the classification score. The pioneering work of the CNN-based two-stage framework is MDNet \cite{MDNet}. Based on MDNet, extensions include meta learning \cite{metatracker}, adversarial learning \cite{VITAL}, online regularization \cite{branchout}, and reciprocative learning \cite{DATNIPS}. Although much progress has been made, these trackers still suffer from the over-fitting problem due to limited online samples, which severely impedes the great potential of CNNs for visual tracking. In this paper, we extend the two-stage framework by incorporating our AH, which effectively mimics the human imaginary mechanism to hallucinate novel positive samples for training a more robust CNN classifier.
\\
\noindent\textbf{Visual tracking by data augmentation.} 
To facilitate the learning of few online training samples in visual tracking, several data augmentation based trackers have been proposed, including UPDT \cite{UPDT}, SINT++ \cite{SINT++} and VITAL \cite{VITAL}. More specifically, UPDT uses several simple data augmentation techniques (e.g., shift, blur and rotation) for correlation filter tracking. SINT++ and VITAL respectively employ deep reinforcement learning and adversarial learning to generate hard positive samples that are occluded by the randomly generated rectangular masks. However, they only use well-designed geometric transforms or fixed shapes of occlusion masks to generate plausible samples. There exists a big gap between the generated samples and real deformable samples. In comparison, the proposed AH learns various non-linear deformations from real instance pairs for transfer.
\\
\noindent\textbf{Augmentation-based few-shot learning.}  
Recently, several augmentation-based few-shot learning methods have been proposed. In \cite{dataauggan} and \cite{hallucinator}, a data augmentation generative adversarial network and a multilayer perceptron are respectively proposed to randomly generate additional samples. Based on the assumption that the deformation between two same-class instances is linear, the authors in \cite{hallucinating} train their generator in a supervised manner. \cite{deltaencoder} improves the above assumption and proposes an encoder-decoder network to learn deformations in a self-reconstruction manner. However, such a way may lead to the domain shift problem. In this work, we overcome this problem by applying the learned deformations between a pair of same-identity instances to instances with other identities via adversarial learning, which makes AH generalize better to unseen instances, even for low-quality online tracked instances.

\section{Proposed Method}
In this section, we firstly introduce the proposed adversarial hallucinator (AH), and then propose the selective deformation transfer (SDT) method for better hallucination. Finally, we detail our hallucinated adversarial tracker (HAT).

\subsection{The Adversarial Hallucinator}
The goal of this paper is to enable the CNN-based trackers to have the capability of ``imagination" or ``hallucination" like humans. Inspired by several recent works \cite{deltaencoder,hallucinator} in few-shot learning, we propose a novel adversarial approach, namely adversarial ``hallucinator" (AH), which consists of an encoder and a decoder, to hallucinate diverse positive samples for visual tracking. The encoder in AH learns to extract non-linear transferable deformations between pairs of same-identity instances, while the decoder applies these deformations to an unseen instance in order to generate diverse reasonable deformable samples of the instance. 

To effectively train AH, we collect a large number of pairs of same-identity instances from the snippets in the ImageNet-VID \cite{ILSVRC} dataset. We call this dataset as $\mathbb{D}_T$. For each pair of instances $(x^{a}_{1}$, $x^{a}_{2})$ with the same identity $a$ in $\mathbb{D}_T$, we randomly select another pair of instances $(x^{b}_{1}$, $x^{b}_{2})$ with different identity $b$ to constitute a quadruplet training sample ($x^{a}_{1}$, $x^{a}_{2}$, $x^{b}_{1}$, $x^{b}_{2}$). We collect a large number of such quadruplet training samples to generate a new dataset $\mathbb{D}_Q$. Then we use the pre-trained feature extractor $\phi(\cdot)$ to extract deep features of all the instances in $\mathbb{D}_Q$. Therefore, the instances in $\mathbb{D}_Q$ are represented by these pre-computed feature vectors, which are more suitable for training. 

We now use the dataset $\mathbb{D}_Q$ to train the proposed AH (i.e., the generator $G$) comprised of the encoder $E_n$ and the decoder $D_e$. For each quadruplet training sample ($x^{a}_{1}$, $x^{a}_{2}$, $x^{b}_{1}$, $x^{b}_{2}$), we feed the concatenated feature vector $[\phi({x^{a}_{1}}), \phi({x^{a}_{2}})]$ to the encoder $E_n$. Let $z^a=E_{n}([\phi({x^{a}_{1}}), \phi({x^{a}_{2}})])$ be the encoder output, which is a low-dimensional vector and represents a plausible transformation. Then, we apply this transformation to a novel instance $x^b_1$ in order to generate a reasonable deformable sample $\hat{x}^{b}= D_{e}([z^a, \phi(x^{b}_{1})])$. 

To guarantee that the generated sample $\hat{x}^{b}$ has the same identity as the input  $\phi(x^{b}_{1})$, i.e., their data distributions should be similar, we employ a discriminator $D$, which is jointly optimized with our AH in an adversarial manner. The whole optimization process can be formulated as:
\begin{equation} \label{eq::combine}
 \begin{split}
\mathcal{L}_{adv} = \min_{G} \max_{D} \mathbb{E}_{x^{b}_{1}, x^{b}_{2}\sim P_{data}(x^{b}_{1})}\left[\log{D([\phi{(x^{b}_{1})}, \phi{(x^{b}_{2})}])}\right]  \\
+\mathbb{E}_{x^{b}_{1}\sim P_{data}(x^{b}_{1}), \hat{x}^{b} \sim P_{data}(\hat{x}^{b})}\left[\log{(1-D([\phi{(x^{b}_{1})}, \hat{x}^{b}]))}\right],
 \end{split}
\end{equation}
where $G=\{E_n, D_e\}$. The proposed AH tries to minimize the adversarial loss $\mathcal{L}_{adv}$ while $D$ aims to maximize it. That is, AH aims to generate samples that fit $P_{data}(x^{b})$ so that the  discriminator cannot discriminate the real pair ($x^{b}_{1}, x^{b}_{2}$) from the fake pair ($x^{b}_{1}, D_{e}([z^a, \phi(x^{b}_{1})])$). Thus, by optimizing the adversarial loss, the proposed AH can effectively generate samples having the same identity as the input $x^{b}_{1}$. 


 \begin{figure}
\begin{center}
   \includegraphics[width=1.00\linewidth]{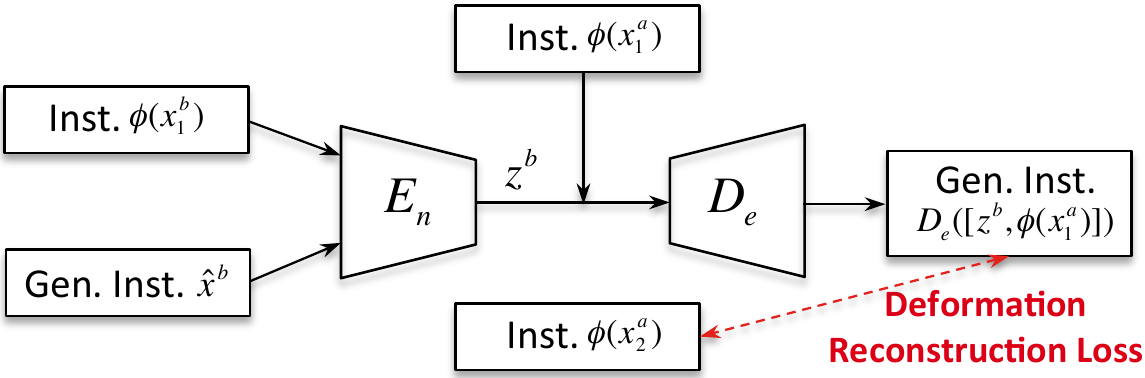} 
\end{center}
\vspace{-0.4cm}
 \caption{Illustration of our deformation reconstruction loss.}
\label{defloss}
\end{figure}

\noindent\textbf{Deformation Reconstruction Loss.} By solely minimizing the adversarial loss to train AH, we can only get  a random generated sample $D_{e}([z^a, \phi(x^{b}_{1})])$ that has the same identity as the input $x^{b}_{1}$. However, we cannot ensure that the generated $\hat{x}^{b}$ effectively learns the non-linear deformation between $x^a_1$ and $x^a_2$. In \cite{hallucinating}, based on the linear deformation assumption, the authors mine various resulting samples and use these samples to train the network in a standard supervised fashion. But since the actual deformations considered in this paper are non-linear, their method is limited and it is even impossible to mine accurate non-linear transformed samples.


To solve this problem, we present a deformation reconstruction (DR) loss, which can be used to train the proposed AH in a self-supervised manner. Assume that the generated sample $\hat{x}^{b}$ correctly encodes the transformation $z^a$ between $x^{a}_{1}$ and $x^{a}_{2}$. Then, the original sample $x^{a}_{2}$ should also be reconstructed by applying the transformation between ${x}^{b}_{1}$ and the generated $\hat{x}^{b}$ to $x^{a}_{1}$. Therefore, we define a DR loss, which can be described as: 
\begin{equation}\label{eq::mse}
\mathcal{L}_{def} = ||D_{e}([{z}^b, \phi(x^{a}_{1})]) - \phi(x^{a}_{2})||_{2}, \\
\end{equation}
where ${z}^b=E_{n}([\phi({x^{b}_{1}}), \hat{x}^{b}])$. As illustrated in Fig. \ref{defloss}, our AH can learn how to correctly perform deformation transfer by itself, and we do not need to use additional samples.
The overall loss function of the proposed adversarial hallucinator is written as:
\begin{equation}\label{eq::all}
\mathcal{L}_{overall} = \mathcal{L}_{adv} + \lambda\mathcal{L}_{def}, \\
\end{equation}
where $\lambda$ is a hyper-parameter.

\subsection{Selective Deformation Transfer}
For humans, we can easily learn deformations from different classes of objects, and applying these deformations to a similar or the same class of objects is easier to be done than applying them to a totally different class of objects. For example, we learn different pose variations from one person, and then those variations are more reasonable and easier to be transferred to another person rather than a car. Based on this observation, we present our selective deformation transfer (SDT) method to better select deformations which are more suitable for transfer.

Let we denote the training dataset $\mathbb{D}_T$ and the online tracking videos as the source domain and the target domain, respectively. In our SDT method, we do not use all the pairs of instances $(x^{a}_{1}$, $x^{a}_{2})$ from the source domain to perform data augmentation on the initial target exemplar $x^e$ of a given video in the target domain. Instead, for each $x^e$, we search a certain number of snippets with similar high-level or semantic characteristics from the source domain. We only use the instances in the snippets, according to the searching results, to perform data augmentation in the online tracking process. 
 \begin{figure}
\begin{center}
   \includegraphics[width=1.00\linewidth]{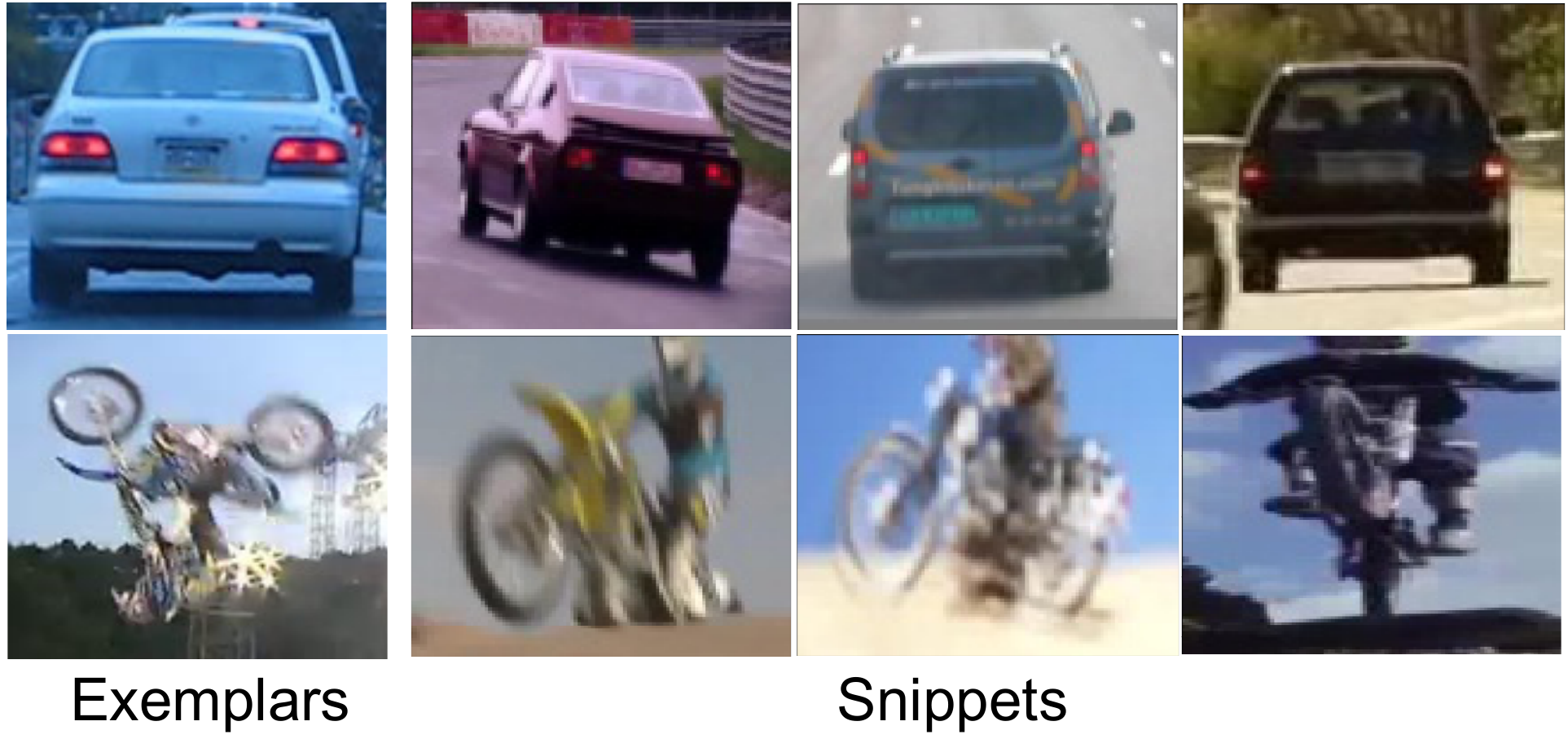} 
\end{center}
\vspace{-0.4cm}
 \caption{Snippets in the source domain that have similar semantic characteristics with target exemplars. The 1st column shows target exemplars from online test videos. The 2nd to 4th columns indicate the corresponding 1st, 2nd and 3rd nearest snippets in the source domain. Each snippet is visualized using its first instance.}
\label{retrieval}
\end{figure}

\noindent\textbf{Snippet descriptor}. Let $N_s$ be the number of snippets in the source domain $\mathbb{D}_T$. We describe each snippet descriptor in the source domain as $\{\psi{(s_i)}\}^{N_s}_{i=1}$, where $\psi{(\cdot)}$ represents a descriptor calculation function and $s_i$ is the snippet with the identity $i$. Moreover, $s_i$ can be further described as $s_i = \{x^i_{j}\}_{j=1}^{N_{s_{i}}}$, where $N_{s_{i}}$ is the number of instances in the snippet $s_i$. In order to search for suitable snippets, an essential problem is how to effectively calculate snippet descriptors. Since our goal is to search instances in the snippets that have similar or the same classes with the target exemplar $x^e$, we use the deep features extracted from the very deep convolutional layers, which provide rich high-level or semantic information, to calculate each snippet descriptor $\psi{(s_i)}$:
\begin{equation}\label{eq::descriptor}
\psi{(s_i)} = \frac{1}{N_{s_{i}}} \sum \nolimits_{j=1}^{N_{s_{i}}}\varphi{(x^i_j)}, \\
\end{equation}
where $\varphi{(\cdot)}$ denotes the convolutional feature extractor. Note that almost any existing deep CNNs can be used as our feature extractor $\varphi{(\cdot)}$. In this work, we use the pre-trained ResNet34 \cite{resnet} model and remove the last fully-connected layer for feature extraction. 
\\
\noindent\textbf{Nearest neighbor ranking}.
After obtaining snippet descriptors $\{\psi{(s_i)}\}^{N_v}_{i=1}$, for the target exemplar descriptor $\varphi{(x^e)}$ in a test video, we search for its nearest-neighbor snippets in the source domain. Specifically, we calculate the Euclidean distance between $\varphi{(x^e)}$ and each snippet descriptor, and rank the snippets in the ascending order. We select the top $T$ snippets $\{s_j\}_{j=1}^{T}$, and collect pairs of same-identity instances from the selected snippets into a dataset $\mathbb{D}_S$ for transfer. In order to reduce the retrieval time, we perform feature extraction for all the instances in the source domain in an offline manner, and calculate the snippet descriptors in advance. Note that there are about $6,500$ snippets in the source domain, and the whole retrieval step can be implemented within 3 seconds. 

Fig. \ref{retrieval} shows that the proposed SDT method can effectively select the snippets that have similar semantic characteristics with the target exemplar into $\mathbb{D}_S$ for better hallucination. 

 \begin{figure}
 \vspace{-0.1cm}
\begin{center}
   \includegraphics[width=1.00\linewidth]{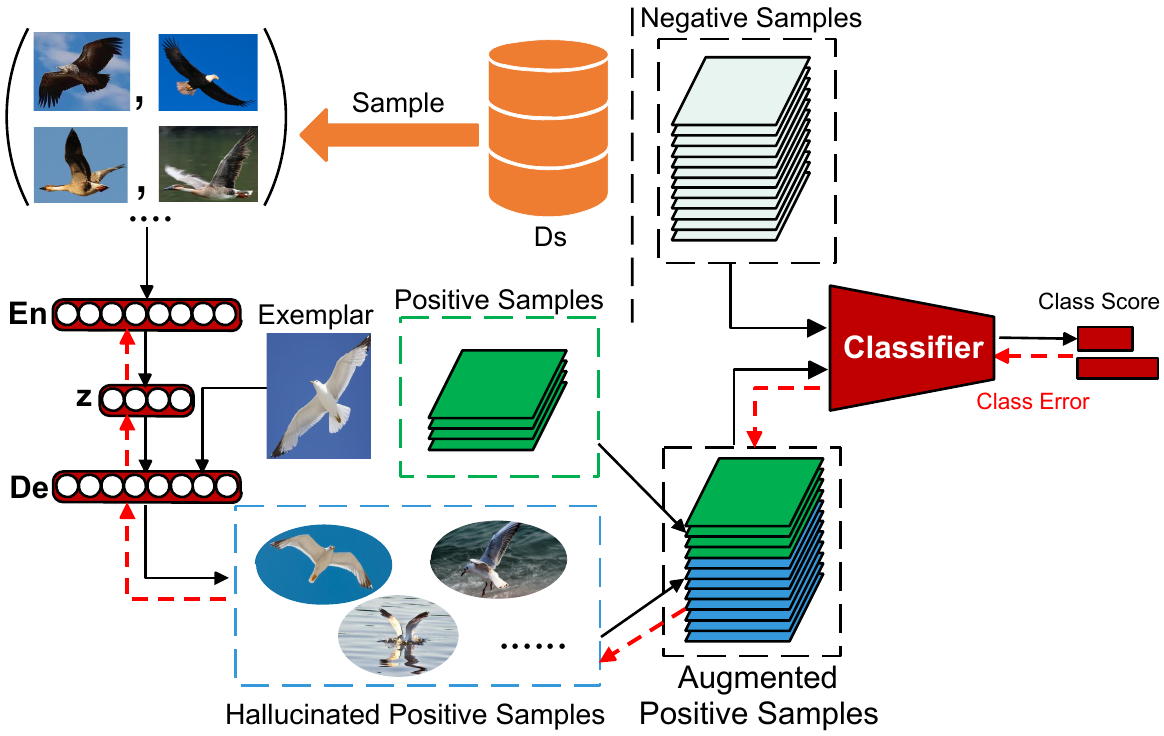} 
\end{center}
\vspace{-0.4cm}
 \caption{Online tracking with hallucination. Given the target exemplar of a test video, we use AH to learn non-linear deformations from pairs of instances in $D_S$ and hallucinate diverse positive samples based on the exemplar. We create an augmented positive set by adding the hallucinated positive samples. The online classifier is jointly optimized with AH. The black and dotted red arrows respectively indicate the forward and back-propagation steps. }
\label{pipeline}
\vspace{-0.1cm}
\end{figure}
\subsection{Proposed Tracking Algorithm}
After training the proposed AH in an end-to-end offline manner and obtaining the selective dataset $\mathbb{D}_S$, we illustrate how we perform hallucinated tracking in an online tracking-by-detection framework. The details of three main components of HAT are given as follows:
\\
\textbf{Joint model initialization.} Given the initial target exemplar $x^e$ in the first frame, we randomly draw 32 positive samples and 96 negative samples around it as in \cite{MDNet} in each iteration. Since these samples are highly spatially overlapped, they cannot capture rich appearance variations. Thus directly using these samples to train the network may lead to the over-fitting problem. To alleviate this problem, we randomly select various pairs of instances in the selective dataset $\mathbb{D}_S$, and use the proposed AH to learn reasonable deformations in the pairs, and then apply those deformations to the target exemplar $x^e$ in order to generate diverse deformable target samples, which are labeled as positive. As illustrated in Fig. \ref{pipeline}, we use both the augmented positive samples and negative samples to jointly update AH and the fully-connected layers in the classifier for $N_1$  iterations.
\\
\textbf{Online detection.} Given an input frame, we first randomly draw samples around the target location estimated in the previous frame. Then, we feed all these samples to the classifier in order to select the best candidate with the highest classification score. Finally, we refine the final target location using the bounding box regression as in \cite{MDNet}.
\\
\textbf{Joint model update.} The joint model update step is similar to the joint model initialization step. First, in each frame, we randomly draw positive and negative samples around the estimated target location. Second, we perform data augmentation to the target exemplar $x^e$ using the proposed AH as described in the joint model initialization step. Finally, we use all the samples to jointly update the fully-connected layers in the classifier and the proposed AH for $N_2$ iterations.

Note that our AH learns generic deformations during the offline learning step. Based on the well-learned offline model, AH can effectively adapt to the online specific deformations of the tracked instance by jointly updating with the classifier.

Jointly training AH and the online classifier has two main benefits. First, there still exists the domain gap between the offline training data and the online tracking data, our joint learning schema makes allowances for errors made by  AH due to the domain gap. Second, the joint learning facilitates AH to generate diverse complement positive samples that are more useful for classification, which helps the classifier generalize well in the low-data tracking regime. 



\renewcommand\arraystretch{1.3}
\begin{table}[tp]  
\setlength{\abovecaptionskip}{0.15cm}
  \newcommand{\tabincell}[2]{\centering\begin{tabular}{@{}#1@{}}#2\end{tabular}}
  \centering
 \fontsize{9.2}{8}\selectfont   
  \caption{DPRs (\%) and AUCs (\%) obtained by the variations of the proposed HAT and  the baseline tracker on the OTB-2013, OTB-2015 and OTB-50 datasets. $r$ represents the ratio of positive and negative training samples. The \textbf{{\color{red}red bold}} fonts and \emph{{\color{blue}blue italic}} fonts respectively indicate the best and the second best results.}  
  \label{ablation}
    \centering\begin{tabular}{|c|c|c|c|c|c|c|}  
    \hline  
    \multirow{2}{*}{Trackers}&  
    \multicolumn{2}{c|}{OTB-2013}&\multicolumn{2}{c|}{OTB-2015}&\multicolumn{2}{c|}{ OTB-50}\cr\cline{2-7}  
    &DPR&AUC&DPR&AUC&DPR&AUC\cr  
    \hline  
    \hline  
    \tabincell{c}{Base-MDNet\\($r=1/3$)}&90.9&66.8&87.3&64.3&82.2&58.6\cr\hline  
    \hline  
   \tabincell{c}{\textbf{HAT}\\\textbf{($r=2/3$)}}&91.4&67.4&90.2&66.1&86.7&61.6\cr\hline  
     \tabincell{c}{\textbf{SDT-HAT}\\ \textbf{($r=2/3$)}}&92.6&68.6&90.3&\emph{{\color{blue}66.5}}&87.2&62.0\cr\hline  
   \tabincell{c}{\textbf{HAT}\\ \textbf{($r=1/1$)}}&\emph{{\color{blue}93.2}}&\emph{{\color{blue}68.7}}&\emph{{\color{blue}90.8}}&66.3&\emph{{\color{blue}87.9}}&\emph{{\color{blue}62.3}}\cr\hline  
   \tabincell{c}{\textbf{SDT-HAT}\\ \textbf{($r=1/1$)}}& \textbf{{\color{red}95.1}}&\textbf{{\color{red}69.6}}&\textbf{{\color{red}91.6}}&\textbf{{\color{red}66.9}}&\textbf{{\color{red}89.4}}&\textbf{{\color{red}63.2}}\cr\hline  
   \end{tabular}  
\end{table}  

\renewcommand\arraystretch{1.3}
\begin{table}[tp]  
\setlength{\abovecaptionskip}{0.15cm}
  \newcommand{\tabincell}[2]{\centering\begin{tabular}{@{}#1@{}}#2\end{tabular}}
  \centering
  \vspace{0.5cm}
 \fontsize{9.2}{8}\selectfont   
  \caption{DPRs (\%) and AUCs (\%) obtained by our HAT equipped with an AH with (HAT-w-Up) and without (HAT-w/o-Up) online update  on the OTB-2013, OTB-2015 and OTB-50 datasets.}  
  \label{ablation_update}
    \centering\begin{tabular}{ccccccc}  
    \hline  
    \multirow{2}{*}{Trackers}&  
    \multicolumn{2}{c}{OTB-2013}&\multicolumn{2}{c}{OTB-2015}&\multicolumn{2}{c}{ OTB-50}\cr\cline{2-7}  
    &DPR&AUC&DPR&AUC&DPR&AUC\cr  
    \hline  
   \tabincell{c}{HAT-w/o-Up}& 91.8&68.0&90.0&66.4&86.6&62.0\cr\hline  
   \tabincell{c}{HAT-w-Up}& 95.1&69.6&91.6&66.9&89.4&63.2\cr\hline  
   \end{tabular}  
\end{table}  

\section{Experiments}
\subsection{Implementation Details}
\noindent \textbf{Network architecture and training.} We use the original network architecture in MDNet \cite{MDNet} as our backbone network, which consists of three convolutional layers and three randomly initialized fully-connected layers. Since HAT does not need to perform the multi-domain learning, the three convolutional layers in HAT share the same weights with the first three convolutional layers in VGG-M \cite{vggm}. The encoder and decoder sub-networks in AH and the discriminator are all designed as three-layer perceptrons with a single hidden layer of 2048 units. The output of the encoder is a 64-dimensional vector. The discriminator output is a probability score. Each layer in these networks is followed by a ReLU activation. Each pair of instances in $\mathbb{D}_T$ is collected from two frames (within the nearest 20 frames) in a snippet from ImageNet-VID.   For the offline training of AH, the hyper-parameter $\lambda$ in Eqn. (\ref{eq::all}) is experimentally set to 0.5. Since the original adversarial loss may lead to unstable training, we use the gradient penalty as in \cite{wgangp} for stable training. We use the Adam solver to optimize both the AH and the discriminator with a learning rate of $2 \times 10^{-4}$ for $5 \times 10^{5}$ iterations. For the joint model initialization step in online tracking, AH is optimized by using the Adam solver with a learning rate of $1.2 \times 10^{-4}$ for $35$ iterations. In the joint model update step, we train our AH for $15$ iterations. For the SDT method, $T$ is set to $2,000$.

We implement our method using PyTorch \cite{pytorch} on a computer with an i7-4.0 GHz CPU and a GeForce GTX 1080 GPU. The average tracking speed is 1.6 FPS. 
\\
\noindent \textbf{Instance representation.} In all the experiments, the instances are represented by pre-computed feature vectors. We use the fixed three convolution layers in the backbone network of the proposed HAT as the feature extractor $\phi(\cdot)$. We first resize all the instances to a fixed size of $107\times107\times3$, and then feed them to the feature extractor in order to extract 4608-dimensional feature vectors.
\\
\noindent \textbf{Evaluation methodology.} For the OTB datasets, we adopt the distance precision (DP) and overlap success (OS) plots for evaluation. We report both the DP rates at the threshold of 20 pixels (DPR) and the Area Under the Curve (AUC). For VOT2016, the expected average overlap (EAO), accuracy, failures and robustness are adopted to evaluate each tracker.

 \begin{figure}
 \vspace{-0.25cm}
\begin{center}
   \includegraphics[width=1.00\linewidth]{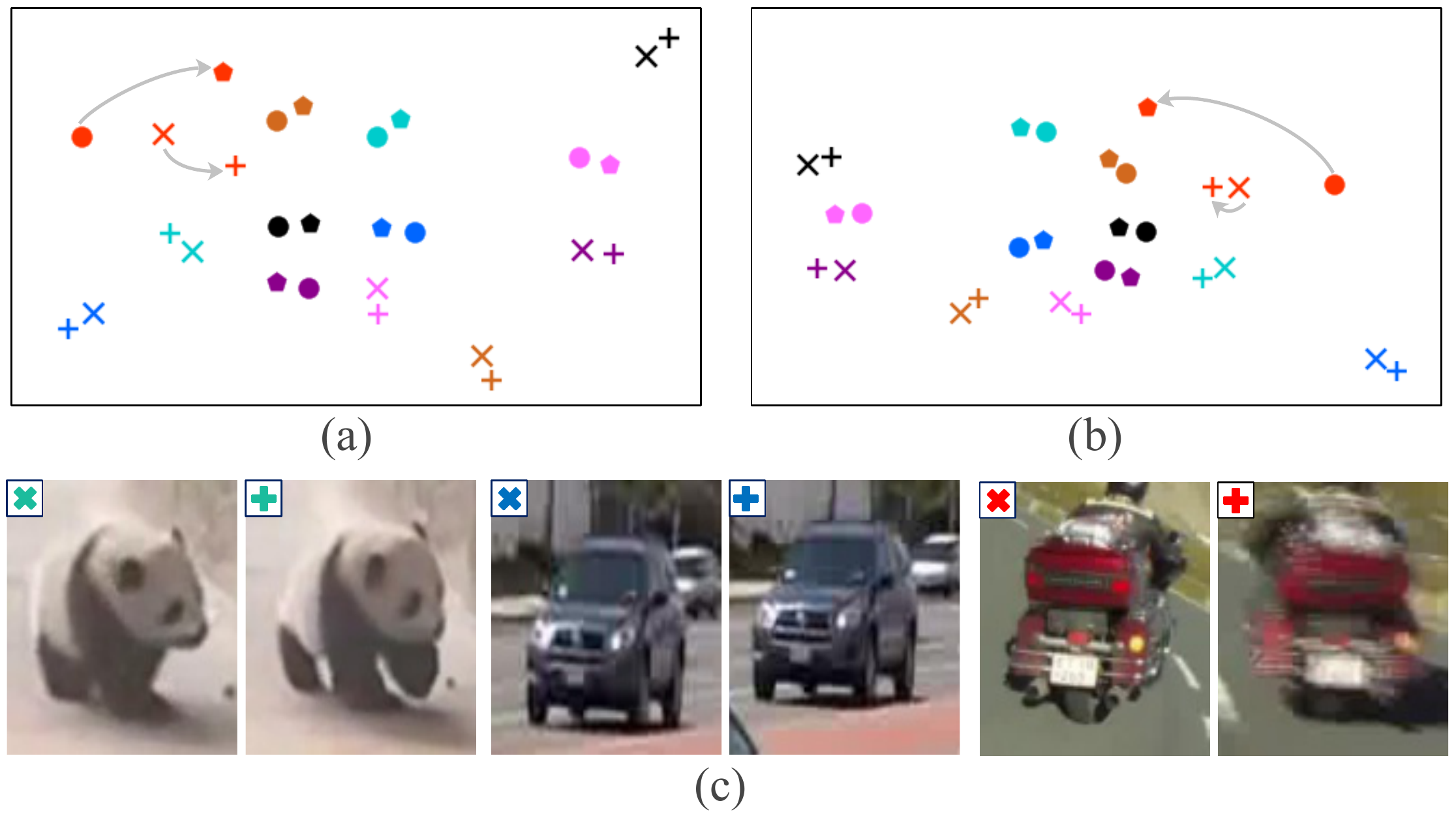} 
\end{center}
\vspace{-0.4cm}
 \caption{(a, b) t-SNE visualizations of the samples hallucinated by the offline learned AHs trained \textbf{(a) with} and \textbf{(b) without the use of the DR loss.} A pair of two instances ($x^{a}_{1}$, $x^{a}_{2}$) used for deformation extraction are shown as a circle and a polygon. By applying the deformation to a novel instance $x^{b}_{1}$ (cross), AH can effectively hallucinate a deformable sample $\hat{x}^{b}$ (plus). Each quadruplet set ($x^{a}_{1}$, $x^{a}_{2}$, $x^{b}_{1}$, $\hat{x}^{b}$) is colored uniquely. \textbf{(c) Hallucinated sample visualization.} The hallucinated samples (plus) in (a) are visualized using their nearest real-instance neighbors in the feature space. }
 
\label{hallucination}
\end{figure}
\subsection{Ablation Study}
First, we analyze the factors that may affect the performance of HAT in Table \ref{ablation}, including the SDT method and the ratio $r$ of positive and negative training samples. Then, we analyze the impact of the online update of AH as shown in Table \ref{ablation_update}. Finally, we analyze the influence of the proposed DR loss and visualize the learned hallucinations in Fig. \ref{hallucination}. Note that our baseline tracker is MDNet, which does not perform the multi-domain learning for fair comparison. Since MDNet uses 32 positive samples and 96 negative samples in each iteration for training, we simply call it as Base-MDNet ($r=1/3$). 

\noindent \textbf{SDT method.} In Table \ref{ablation}, by employing the proposed SDT method, HAT gains the improvements on all the three datasets in terms of both DPR and AUC. Specifically, SDT-HAT ($r=1/1$) achieves the best DPR (95.1\%) by improving 1.9\% of HAT ($r=1/1$) on OTB-2013, which can be explained that the instances selected by our SDT method share more semantic characteristics with the exemplar, thus providing more reasonable transformations for hallucination. 
\\
\noindent \textbf{Ratios of positive and negative training samples.} Since the original ratio of positive and negative samples used in MDNet is $1/3$ (or $32/96$), there still exists the data imbalance problem. We use our AH to hallucinate diverse positive samples such that the ratios of augmented positive samples and negative samples are respectively $2/3$ (or $64/96$) and $1/1$ (or $96/96$). In Table \ref{ablation}, we can find that the tracking performance is significantly improved by adding more hallucinated positive samples for learning. The promising results (DPR: 95.1\%, AUC: 69.6\%) can be achieved by HAT on OTB-2013 when the ratio is set to $1/1$, which demonstrates that the balanced data can lead to better results. 
\\
\noindent \textbf{Impact of online update.} As shown in Table \ref{ablation_update}, we can find that HAT-w-Up significantly outperforms HAT-w/o-Up in all the three datasets in terms of both DPR and AUC metrics. This is because that online update of AH can effective alleviate the domain gap between the offline training data and the online tracking data. In addition, even without online updating AH, HAT-w/o-Up still achieves much better results than the baseline tracker as shown in Table \ref{ablation}, which demonstrates the effectiveness of the proposed AH.
\\
\noindent \textbf{Influence of DR loss.}  We apply t-SNE to visualize the samples hallucinated by the offline learned AHs trained by using our DR loss in Fig. \ref{hallucination}(a) or without using it in Fig. \ref{hallucination}(b). Note that the transformed instances (crosses) are unseen during the offline training. As can be seen, the AH trained by using the DR loss generates the samples that keep more information of the original learned deformations. For example, since the deformation between the two instances (red circle and polygon) in Fig. \ref{hallucination} is relatively large, the generated sample (red plus) in Fig. \ref{hallucination}(a) also lies relatively far away from its exemplar (red cross), which indicates similar large deformation. In comparison, in Fig. \ref{hallucination}(b), without using the DR loss, the learned AH tends to generate random samples (lie close to their exemplars (crosses)), which are irrelevant to the applied deformations.\\ 
\noindent \textbf{Visualizing the hallucinated samples.} We visualize the hallucinated samples in Fig. \ref{hallucination}(c) using their nearest real-instance neighbors in the validation set (including $83,996$ instances), which demonstrates that AH can effectively hallucinate reasonable deformable samples.\\

\begin{figure}
 \begin{center}
 \subfigure{
\begin{minipage}[b]{0.48\linewidth}
  \centerline{\includegraphics[width=4.25cm]{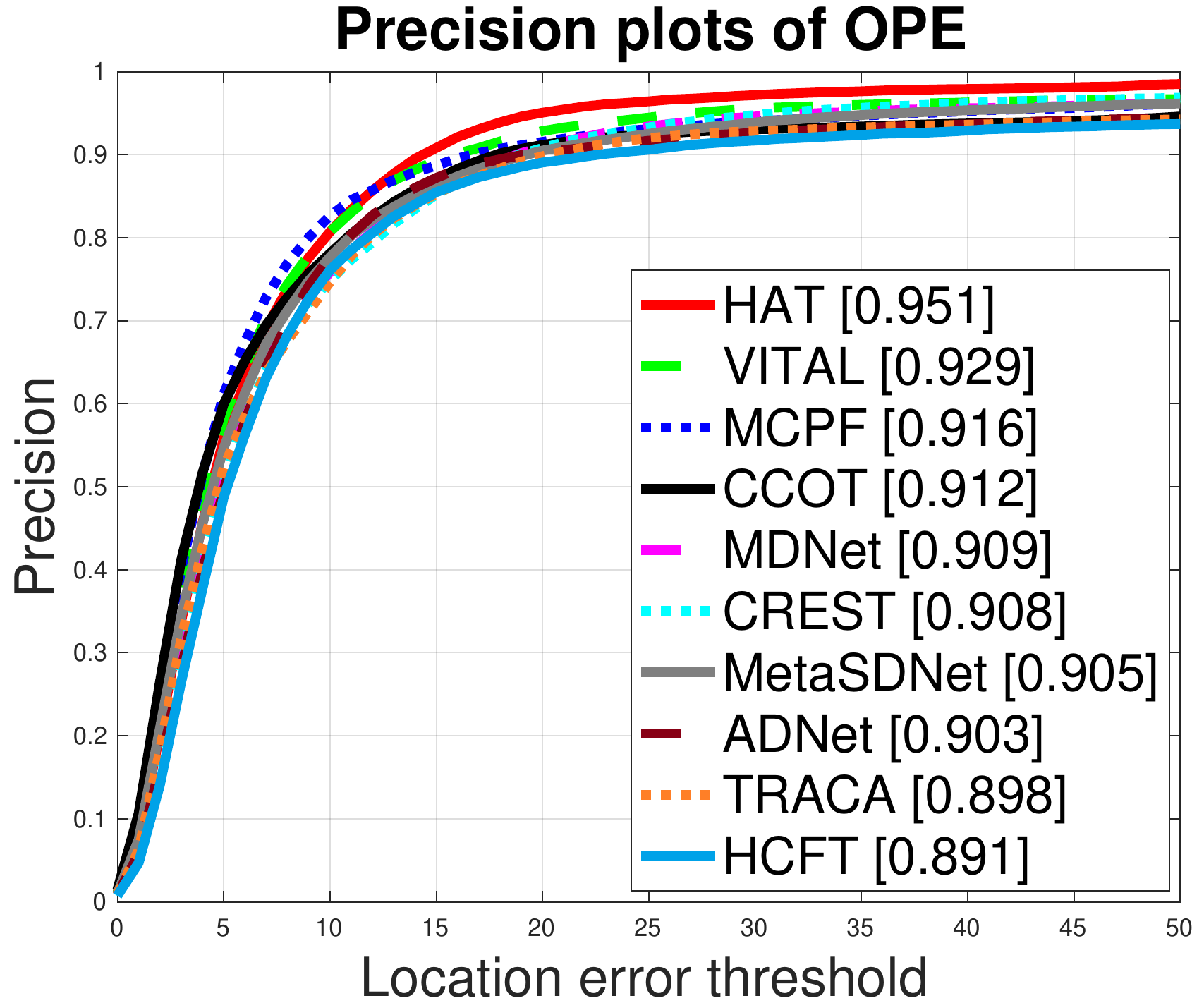}}
\end{minipage}}
\vspace{-0.1cm}
 \subfigure{
\begin{minipage}[b]{.48\linewidth}
  \centerline{\includegraphics[width=4.25cm]{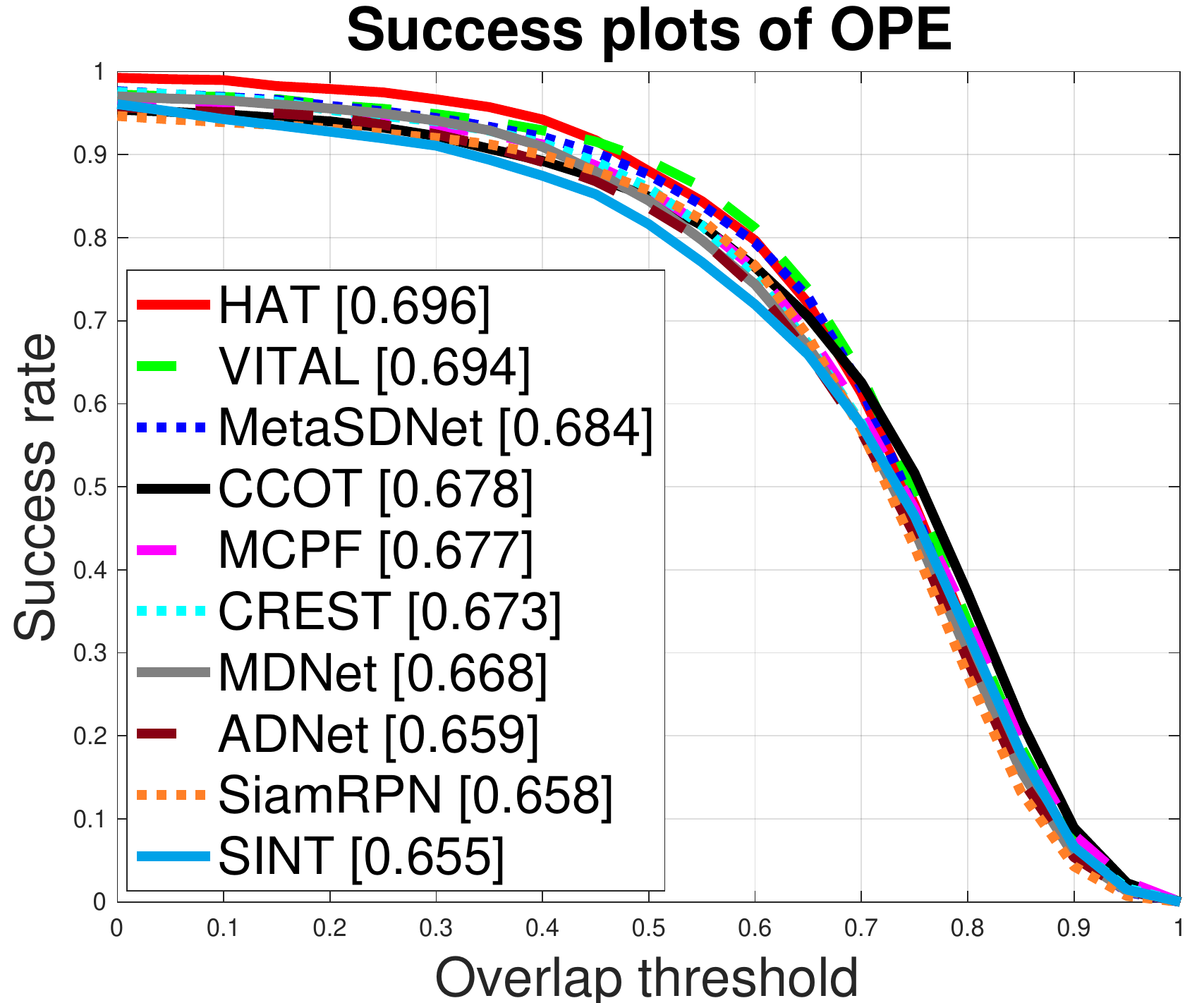}}
\end{minipage}}
\end{center}
\vspace{-0.45cm}
\vfill
   \caption{\label{otb2013} Precision and success plots on OTB-2013 using one pass evaluation.}
\end{figure}
\begin{figure}
 \begin{center}
 \subfigure{
\begin{minipage}[b]{0.48\linewidth}
  \centerline{\includegraphics[width=4.25cm]{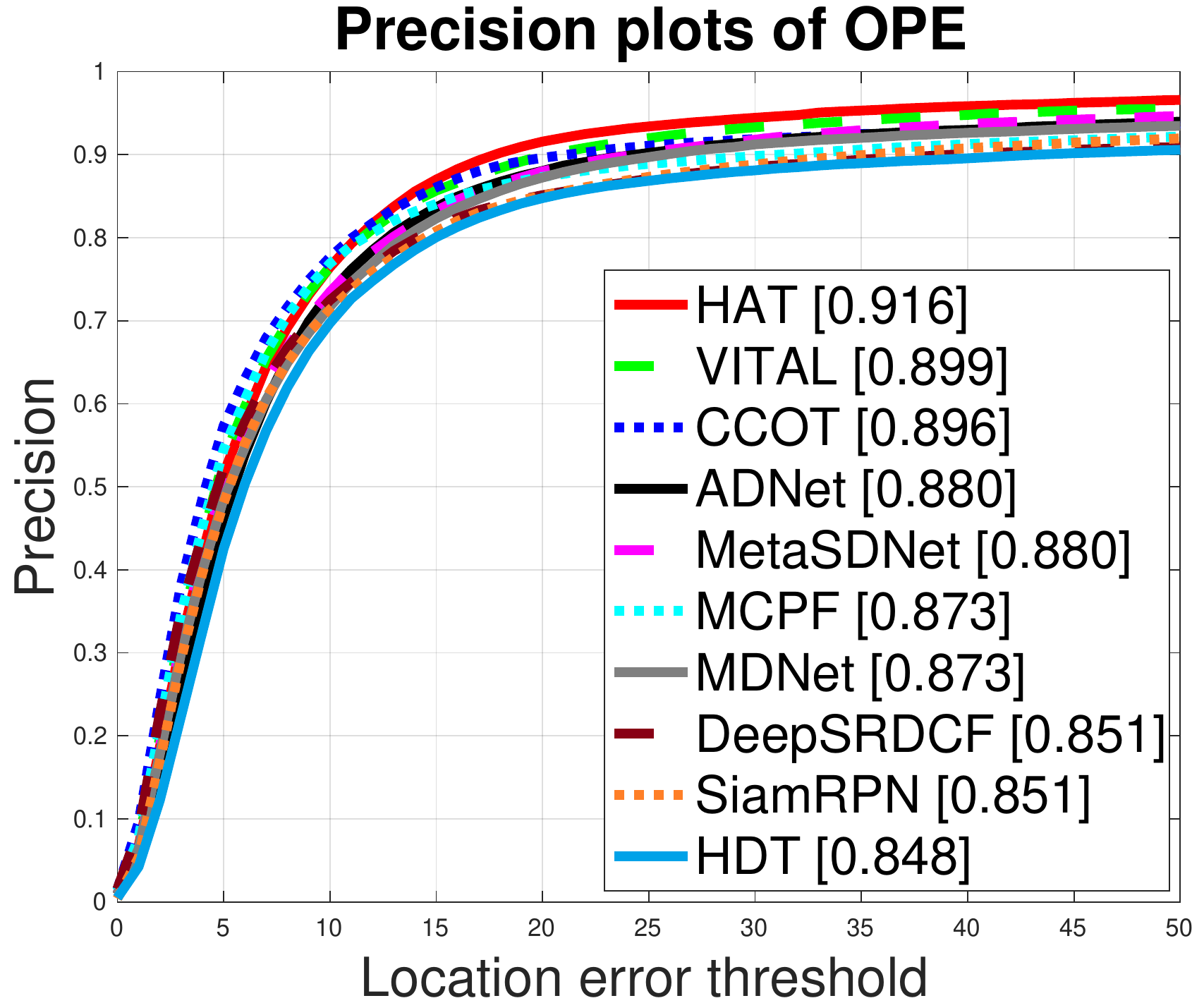}}
\end{minipage}}
\vspace{-0.1cm}
 \subfigure{
\begin{minipage}[b]{.48\linewidth}
  \centerline{\includegraphics[width=4.25cm]{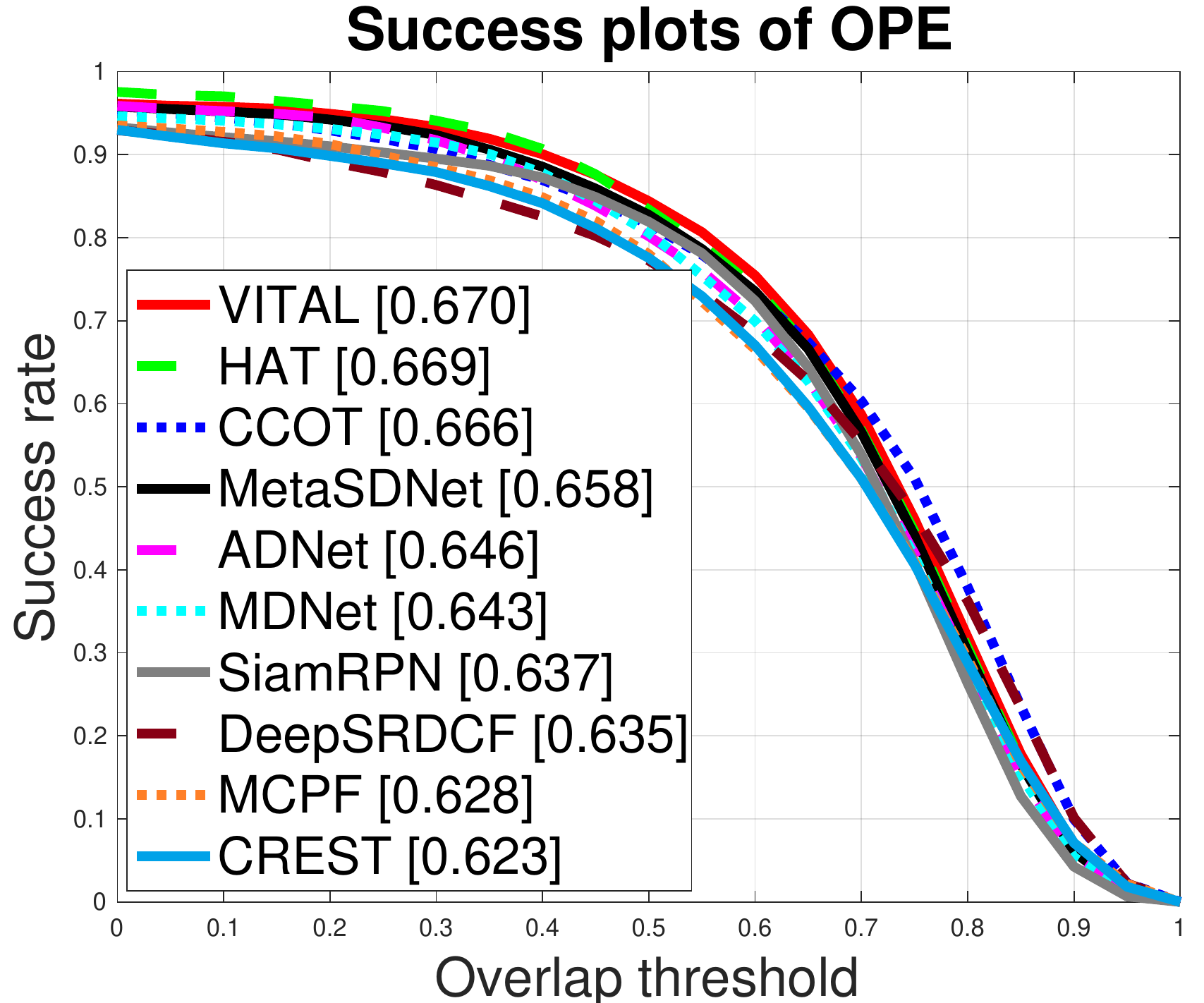}}
\end{minipage}}
\end{center}
\vspace{-0.45cm}
\vfill
   \caption{\label{otb2015}Precision and success plots on OTB-2015 using one pass evaluation.}
\vspace{-0.15cm}
\end{figure}

\subsection{Evaluations on OTB-2013 and OTB-2015}
We compare HAT with 13 state-of-the-art trackers including MDNet \cite{MDNet}, VITAL \cite{VITAL}, MetaSDNet \cite{metatracker}, ADNet \cite{ADNet}, SiamRPN \cite{siamrpn},  CCOT \cite{CCOT}, SiamFC \cite{SiamFC}, TRACA \cite{TRACA}, MCPF \cite{MCPF}, CREST \cite{CREST}, HDT \cite{HDT}, HCFT \cite{HCF} and DeepSRDCF \cite{DeepSRDCF}. For fair comparison, we do not apply the multi-domain learning for MDNet. We only report the top 10 trackers for presentation clarity. 

\renewcommand\arraystretch{1.2}
\begin{table}
\setlength{\abovecaptionskip}{0.15cm}
\newcommand{\tabincell}[2]{\begin{tabular}{@{}#1@{}}#2\end{tabular}}
\begin{center}
\vspace{0.7cm}
\caption {\label{vot16}The EAO, accuracy (Acc.), failures (Fai.) and robustness (Rob.) obtained by HAT and five state-of-the-art trackers on VOT2016. The best and the second best results are highlighted by the \textbf{{\color{red}red bold}} and  \emph{{\color{blue}blue italic}} fonts, respectively.}
\setlength{\tabcolsep}{1.05mm}{
\begin{tabular}{|c|c|c|c|c|c|c|}
\hline
 &  {\tabincell{c}{HAT}}& {\tabincell{c}{MetaSDNet}} & \tabincell{c}{CCOT} &  \tabincell{c}{VITAL} & \tabincell{c}{MDNet}  & \tabincell{c}{Staple} \\
\hline
\multirow{1}{*}{EAO}
  & \emph{{\color{blue}0.32}}& 0.31 &  \textbf{{\color{red}0.33}} & \emph{{\color{blue}0.32}} & 0.26 & 0.30   \\
\hline
\multirow{1}{*}{Acc.}
 & \textbf{{\color{red}0.58}}& 0.54 & 0.54 &  \emph{{\color{blue}0.56}} & 0.54 & 0.55 \\
 \hline
   \multirow{1}{*}{Fai.}
 & \textbf{{\color{red}16.52}}& 17.36 &\emph{{\color{blue}16.58}} &18.37 & 21.08 &23.90 \\
  \hline
 \multirow{1}{*}{Rob.}
 &0.27&\emph{{\color{blue}0.26}}& \textbf{{\color{red}0.24}} & 0.27 & 0.34 & 0.38 \\
\hline
\end{tabular}}
\vspace{-0.4cm}
\end{center}
\end{table}
Fig. \ref{otb2013} shows the results achieved by all the trackers on the OTB-2013 dataset. More particularly, the DPR obtained by HAT is 95.1\%, which is the leading accuracy on the OTB-2013 dataset. Furthermore, HAT achieves the best  AUC (69.6\%) among all the compared trackers, outperforming its baseline tracker MDNet with a large margin of 2.8\%. Compared with the VITAL tracker, the DPR and AUC obtained by our HAT are both higher than those obtained by VITAL on OTB-2013, which empirically shows the superiority of the proposed hallucination method.
In Fig. \ref{otb2015}, HAT achieves the best accuracy (91.6\%) on OTB-2015 followed by VITAL (89.9\%), CCOT (89.6\%) and ADNet (88.0\%). This comparison shows the highest accuracy achieved by HAT among the state-of-the-art deep trackers. In addition, HAT obtains the comparable AUC (66.9\%) to VITAL (67.0\%) on OTB-2015, and it is better than the others. Overall, compared with the 13 state-of-the-art trackers, HAT achieves the better overall performance on the OTB-2013 and OTB-2015 datasets. 



\subsection{Evaluation on VOT-2016}
The proposed HAT is evaluated on VOT-2016 \cite{VOT2016} with the comparison to the state-of-the-art trackers including VITAL \cite{VITAL}, MetaSDNet \cite{metatracker}, CCOT \cite{CCOT}, MDNet \cite{MDNet} and Staple \cite{Staple}. 

As can be seen from Table \ref{vot16}, the proposed HAT and VITAL achieve the comparable EAO (0.32) to CCOT (0.33), and HAT significantly outperforms the baseline tracker MDNet (0.26) with a relative gain of 23\%, which demonstrates the overall effectiveness of the proposed tracker. In terms of accuracy, HAT obtains a relative gain of 4\% compared to VITAL, meanwhile outperforming the other trackers with large margins. Furthermore, our tracker achieves the best results (16.52) in terms of failures, which indicates its stability in visual tracking. In terms of robustness, HAT and VITAL achieves the comparable robustness (0.27) to MetaSDNet (0.26). Meanwhile, HAT outperforms MetaSDNet on the other metrics, including EAO, accuracy and failures.

\section{Conclusions and Future Work}

Inspired by the human imaginary mechanism, we propose an adversarial hallucinator (AH) for data augmentation. A novel deformation reconstruction loss is introduced to train AH in a self-supervised manner. By incorporating AH into a tracking-by-detection framework, the hallucinated adversarial tracker (HAT) is proposed. HAT jointly optimizes AH with the classifier (MDNet) in an end-to-end manner, which facilitates AH to generate diverse positive samples for reducing tracking failures. In addition, we present a novel selective deformation transfer method to further improve the tracking performance. Experiments on three popular datasets demonstrate that HAT achieves the state-of-the-art performance. Except for visual tracking, we believe that our generic AH can be utilized in a variety of tasks, e.g., few-shot and semi-supervised learning. We leave these directions for future work.
\bibliographystyle{named}
\bibliography{ijcai19}

\end{document}